\documentclass{article}

\usepackage{arxiv}

\usepackage[utf8]{inputenc} 
\usepackage[T1]{fontenc}    
\usepackage{hyperref}       
\usepackage{url}            
\usepackage{booktabs}       
\usepackage{amsfonts}       
\usepackage{nicefrac}       
\usepackage{microtype}      
\usepackage{lipsum}		
\usepackage{graphicx}
\usepackage{natbib}
\usepackage{doi}

\usepackage{multirow}
\usepackage{amsmath}
\usepackage{mathtools}
\usepackage{caption}
\usepackage{subcaption}

\title{DeepSet SimCLR: Self-supervised deep sets for improved pathology representation learning}

\author{ \href{https://orcid.org/0000-0003-2822-7146}{\includegraphics[scale=0.06]{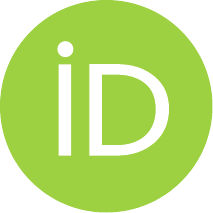}\hspace{1mm}David ~Torpey}\thanks{Alternative email address: 674425@students.wits.ac.za.} \\
	School of Computer Science and Applied Mathematics\\
	University of the Witwatersrand, Johannesburg\\
	South Africa \\
	\texttt{torpey.david93@gmail.com} \\
	\And
	\href{https://orcid.org/0000-0003-0783-2072}{\includegraphics[scale=0.06]{orcid.pdf}\hspace{1mm}Richard ~Klein} \\
	School of Computer Science and Applied Mathematics\\
	University of the Witwatersrand, Johannesburg\\
	South Africa \\
	\texttt{kleinric@gmail.com} \\
}

\renewcommand{\shorttitle}{DeepSet SimCLR}

\hypersetup{
pdftitle={DeepSet SimCLR: Self-supervised deep sets for improved pathology representation learning},
pdfsubject={cs.CV},
pdfauthor={David ~Torpey, Richard ~Klein},
pdfkeywords={Deep learning, Self-supervised learning, Representation learning, Affine transformation, Medical image analysis},
}

\begin{document}
\maketitle

\begin{abstract}
Often, applications of self-supervised learning to 3D medical data opt to use 3D variants of successful 2D network architectures. Although promising approaches, they are significantly more computationally demanding to train, and thus reduce the widespread applicability of these methods away from those with modest computational resources. Thus, in this paper, we aim to improve standard 2D SSL algorithms by modelling the inherent 3D nature of these datasets implicitly. We propose two variants that build upon a strong baseline model and show that both of these variants often outperform the baseline in a variety of downstream tasks. Importantly, in contrast to previous works in both 2D and 3D approaches for 3D medical data, both of our proposals introduce negligible additional overhead over the baseline, improving the democratisation of these approaches for medical applications.
\end{abstract}


\section{Introduction}
Self-supervised learning (SSL) has pushed forward the capabilities of large-scale visual representation learning due to its ability to model large datasets without manual or human annotation \citep{simclr,simclrv2,dino,mocov2}. Instead, these methods derive a supervision signal from the data itself and often are architected to solve a proxy task of some type. Through recent improvements, self-supervised pretraining has been shown to outperform supervised pretraining on many downstream tasks \citep{sslanalysis,simclrv2}. This is particularly important for medical domains, where annotation is extremely expensive, laborious, and time-consuming to obtain. However, these methods, even in their standard 2D setting, are often notably computationally intensive to train, often requiring tens of GPUs to train \citep{dino,barlow_twins,byol,swav} and sometimes requiring hundreds \citep{simclrv2}. 3D extensions of these techniques are even more computationally intensive. It should be noted that even tens of GPUs severely reduce the potential widespread applicability of these techniques.

Motivated by the success of the DeepSet \citep{deepset} architecture for the application of whole slide image (WSI) classification \citep{deepset_wsi}, we propose two variants that build upon a strong modern contrastive SSL baseline. The variants use an intra-scan sampling strategy, with one variant employing the DeepSet architecture, to implicitly model the 3D nature of medical data using an inherently 2D architecture. Whereas previous approaches focus on modelling the 3D volumes using either 3D native architectures \citep{3d_ssl_methods_for_medical_imaging,3d_swin} or computationally-intensive 2D architectures \citep{2d3djointmedicalssl}, we instead focus on making this process more computationally efficient using an architecture proven to work well for the modelling of pathology data \citep{deepset_wsi} while inducing limited computational overhead.

We propose two novel variants of a multi-view contrastive baseline: SimCLR \citep{simclr}. First, we devise a simple and effective slice sampling strategy for the generation of random views. Next, we propose a new architecture where each `view' is a set of images rather than a single image, which is operated upon using a Deep Set network. Our adaptations introduce negligible additional computational cost to the architecture (at most a small multi-layer perceptron). In this way, we inherit the plethora of benefits from the base SimCLR architecture, while improving performance in the medical domain.

Through extensive experiments on various downstream datasets and task types, we show that the two variants consistently outperform the baseline, sometimes by a large margin. Further, we perform analyses of core architectural decisions, including the size of the set fed into the Deep Set, the sampling strategy, and how the Deep Set is embedded within SimCLR.

We note that our goal is to improve upon a standard SSL baseline model, and \emph{not to achieve a new state-of-the-art}. This is because the ubiquity of SSL methods means that any improvements upon them can greatly benefit many domains, particularly those tailored for the medical domain due to its exceptionally prohibitive annotation bottleneck. Our focus is on changes to an SSL baseline that are easy to integrate and deploy, while simultaneously robust and performant with negligible performance overhead. This is in stark contrast to existing works for self-supervision on medical data which are typically computationally intensive compared to standard SSL models \citep{2d3djointmedicalssl,3d_ssl_methods_for_medical_imaging,3d_swin}. As such, we note our contributions:

\begin{itemize}
    \item Two proposed models that implicitly model the 3D nature of pathology data while remaining similarly computationally efficient as, and significantly more performant than, the strong baseline.
    \item A detailed study of the effect of the key hyperparameters introduced by our proposed models.
    \item A comprehensive evaluation on a large set of downstream datasets that cover a variety of medical tasks and pathology modalities.
\end{itemize}

The rest of this paper is structured as follows. In Section \ref{sec:related_work}, we describe the relevant literature in both SSL and its applications to the medical domain. Section \ref{sec:methodology}, we describe the baseline model and our two proposed variants in detail. Section \ref{sec:experiments} contains the experiments and analyses for our variants on multiple downstream datasets and tasks. Lastly, Section \ref{sec:conclusion} contains concluding remarks and proposals for future work.

\section{Related Work}
\label{sec:related_work}
\subsection{Self-Supervised Learning}
Modern approaches to SSL typically fall into one of 3 categories: 1) contrastive, 2) non-contrastive, or 3) clustering. Techniques such as SimCLR \citep{simclr,simclrv2}, NNCLR \citep{nnclr}, and MoCo \citep{moco,mocov2} are contrastive approaches. These approaches aim to pull representations of similar `views' (i.e. those sourced from the same image) together while pulling the representations of dissimilar `views' apart. Non-contrastive approaches such as Barlow Twins \citep{barlow_twins} and BYOL \citep{byol} still discriminate between instances but instead optimise for different objectives that are reliant on negative samples such as contrastive losses. For example, Barlow Twins aims to make the cross-correlation matrix between two batches of representations of views to be as similar to the identity matrix as possible. BYOL is a self-distillation approach that aims to train a teacher-student architecture where the goal of the first network is to predict the representation produced by the other.

Clustering approaches such as SwAV \citep{swav} try to circumvent a strong assumption made by contrastive approaches where views from the same image, whatever they may depict, are trained to be represented by the same vector, and views from different images are always trained to be as dissimilar as possible. SwAV therefore attempts to discriminate between clusters of representations instead of individual instance representations, where clusters contain representations of different views of the same image.

In general, these types of SSL approaches rely on generating the views through data augmentation using a stochastic augmentation function which is typically comprised of transformations such as random cropping, horizontal flipping, Gaussian blurring, and colour jittering. This augmentation function must produce a variety of views while maintaining the semantics between views generated from the same image. There is also a trend towards attempting to better understand the properties of these SSL methods, as they as not yet well understood \citep{simsiam_collapse,phd-prletters-viewpoint,understanding_dim_collapse_in_ssl}.

Another class of SSL techniques are pretext tasks where the network is tasked with solving manual proxy tasks usually based on either image restoration or spatial content. These include techniques such as rotation prediction \citep{predicting_image_rotations}, colourisation \citep{colorful_image_colorization}, inpainting \citep{inpainting}, and image jigsaw puzzles \citep{jigsaw}.

\subsection{Applications of SSL to Medical Data}
Since SSL is a promising and scalable approach to modelling large datasets while being annotation-free, attempts have been made to apply it to the medical domain where obtaining annotations is particularly prohibitive to do at scale. Most SSL approaches in the medical domain can be classed as either 2D or 3D - whether they model 2D medical image data, or 3D volumes from sources such as computed tomography (CT) or medical resonance imaging (MRI) scans.

\cite{improve_ssl_pathology} show that domain-aligned pretraining (i.e. pretraining on a medical dataset) consistently outperforms the standard paradigm of ImageNet pretraining for multiple common SSL methods. Further, they propose an altered set of transformations for the data augmentation pipeline to generate views more aligned with the medical domain. These changes include weaker colour jittering, random vertical flips, and a novel stain augmentation (tailored to WSI images). This new set of transformations improves performance over the standard set of augmentations included in most modern SSL techniques \citep{simclr,byol,barlow_twins}.

\cite{3d_ssl_methods_for_medical_imaging} proposes 3D variants of multiple successful 2D SSL algorithms, including CPC \citep{cpc}, RotNet \citep{predicting_image_rotations}, image jigsaw puzzles \citep{jigsaw}, relative patch prediction \citep{contextpred}, and Exemplar networks \citep{exemplarcnn}. These SSL models are naturally extended into 3D due to their reliance on spatial context. They demonstrate their success for multiple downstream medical tasks.

\cite{3d_swin} proposed a novel 3D vision transformer (ViT) \citep{vit} architecture to facilitate SSL on 3D medical volumes. They propose random cutout and rotation as the transformations for the generation of 3D views as they posit that these are more aligned with human anatomy. This SS ViT is tasked with solving a multi-task objective consisting of inpainting, rotation prediction, and contrastive learning.

\cite{2d3djointmedicalssl} attempts to jointly model 2D and 3D medical data to capture both the spatial semantics of the individual 2D images, as well as long-range dependencies between the slices from a 3D volume. The framework is built upon clustering-based SSL algorithms SwAV \citep{swav} and Deep-ClusterV2 \citep{deepercluster}. A transformer \citep{transformer} with a novel deformable self-attention mechanism is added into the base architecture to capture the aforementioned long-range dependencies of 3D volume data. Our proposed method differs from this approach in that we model the 3D volume as a set or implicitly through sampling, instead of as a sequence with a transformer. Further, our proposed models introduce significantly less computational overhead than the additional transformer and its masked embedding prediction task.

\section{Methodology}
\label{sec:methodology}
Consider a dataset of 3D volumes, such as CT or MRI scans. Each scan $S_i$ is comprised of an \emph{ordered} collection of slices: $S_i = \{x_j \in \mathbb{R}^{H \times W \times C}\}_{j=1}^{L_i}$, where $H \times W \times C$ are the spatial dimensions of slice $x_j$ and $L_i$ is the number of slices in scan $S_i$. We define a dataset of $N$ scans $\mathcal{S} \coloneqq \{S_i\}_{i=1}^N$ and a flattened version consisting of all slices from all scans $\mathcal{S}_\texttt{flat} \coloneqq \cup S_i$. Note that $|\mathcal{S}_\texttt{flat}| = \sum_{i=1}^{N} L_i$.

Define an encoder network $f$ and projection head $g$, which are the equivalent functions as defined in SimCLR \citep{simclr}. Finally, we define a (possibly infinite) set of random augmentation functions $\mathcal{A}$, where each $\alpha \in \mathcal{A}$ is a function that applies a pipeline of random transformations to generate views. We use the same base set of transformations as \cite{simclr} in order to construct $\mathcal{A}$.

\subsection{SimCLR Baseline}
Our baseline model is the standard SimCLR approach. A slice is randomly sampled across all the slices from all scans: $x \in \mathcal{S}_\texttt{flat}$. We generate views using two random augmentation functions: $x_1 = \alpha_1(x)$ and $x_2 = \alpha_2(x)$ where $\alpha_1, \alpha_2 \in \mathcal{A}$. These views are then sent through the encoder and projection head to compute latent representations: $z_1 = g(f(x_1))$ and $z_2 = g(f(x_2))$. The contrastive NT-Xent loss is then computed using these representations. In this way, a positive pair of views is always sourced from the same input slice. This precludes the ability to model the 3D nature of the scans as the model never compares regions from different slices within the same scan.

Instead, it attempts to pull the representations of different views of the same slice together in latent space and push the representations of views from different slices (\emph{even if from different slices potentially within the same scan}) apart. It is unclear apriori whether this assumption is detrimental to transfer learning performance, however, it is the standard SSL contrastive setup, and we make novel adaptations to it to build our two proposed variants: \emph{Per-scan SimCLR} and \emph{DeepSet SimCLR}.

\subsection{Per-scan SimCLR}
The intuition for our Per-scan SimCLR (PS-SimCLR) approach is that the sampling strategy of the baseline precludes the model from accounting for the 3D nature of the scans. We address this by proposing a simple and effective novel sampling strategy that attempts to implicitly encode the 3D nature of the scans during the modelling process. In this way, PS-SimCLR changes how random views are generated but keeps the network architecture consistent with the baseline.

Our sampling strategy is controlled by a sampling width $\omega \in (0, 1]$. This parameter controls the width of a contiguous region (i.e. window) of the scan $s_i \in X$ from which slices can be sampled, as well as the amount of shared semantic and mutual information between random views. We randomly choose the start location $j$ of this window from the range $\{0, \cdots, \omega L_i\}$. The end location of the window is then given by $j + \omega L_i$. The positive pair of random views is then generated by sampling two slices from this window of the scan: $x, x' \in \{x_j, \cdots, x_{j + \omega L_i}\}$. The positive view is pair is then defined as $x_1 = \alpha_1(x)$ and $x_2 = \alpha_2(x')$. The rest of the PS-SimCLR pipeline is identical to the baseline: we compute the NT-Xent loss from representations $z_1 = g(f(x_1))$ and $z_2 = g(f(x_2))$. Note that we default to the baseline strategy of generating views if $L_i$ is sufficiently small, i.e. $L_i < T$ for some threshold $T$ (this happens infrequently).

In contrast to the baseline model, PS-SimCLR pulls representations of different views of \emph{any} slice within the same scan together and contrasts them against representations of slices from any other scan. In this way, we implicitly model the 3D nature of the volume by imposing that views of the same volume should be close in latent space, instead of only views of a single slice of a scan needing to be close in latent space.

\subsection{DeepSet SimCLR}
\begin{figure}
	\centering
	\includegraphics[width=.5\textwidth]{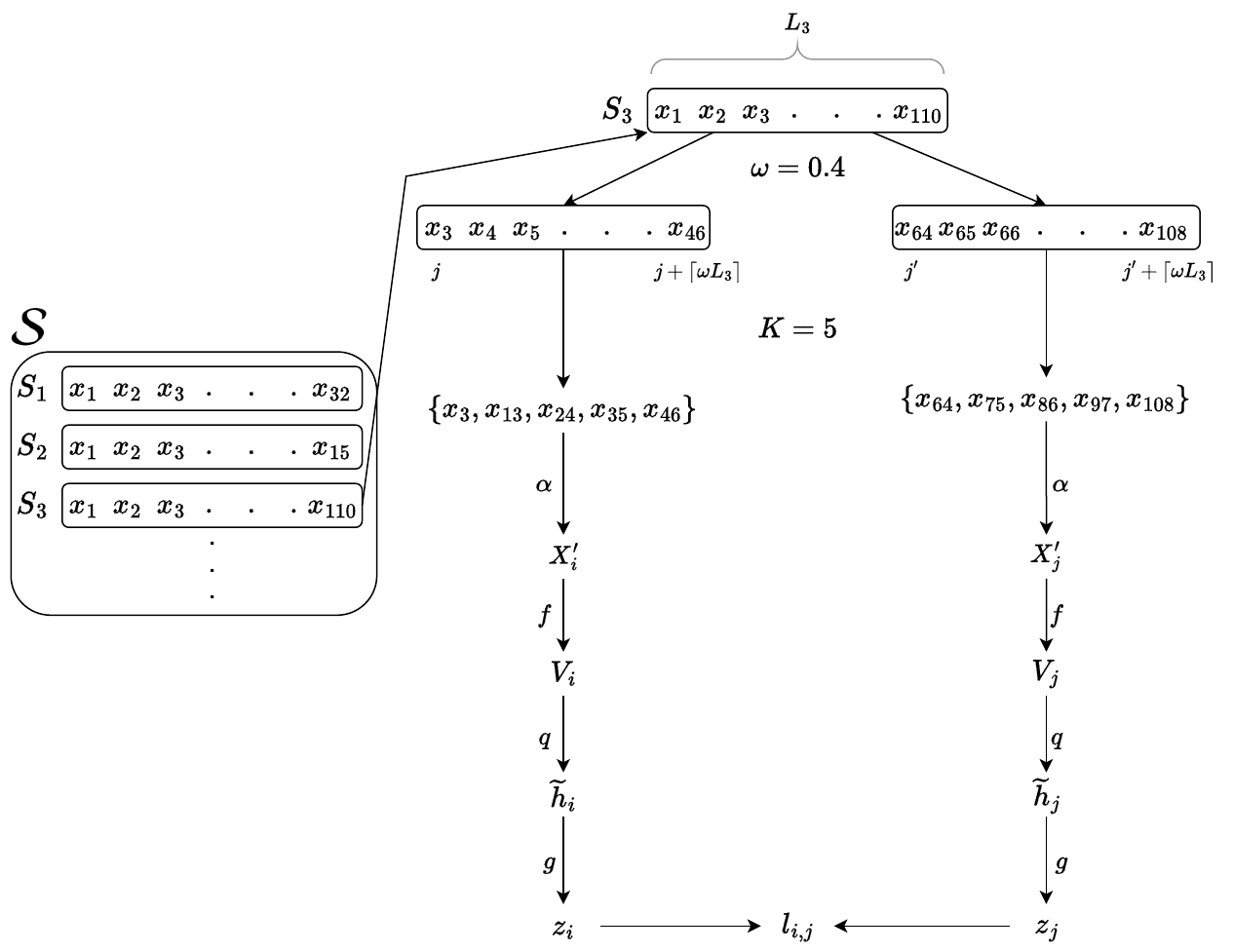}
	\caption{DS-SimCLR architecture.}
	\label{fig:ds_simclr_arch}
\end{figure}

Our DeepSet SimCLR (DS-SimCLR) variant adds a more explicit mechanism to model the 3D nature of the scans in addition to the sampling strategy used in PS-SimCLR. We achieve this by modelling a random view as a set of slices instead of as an individual slice (as in PS-SimCLR and the baseline). Motivated by the success of the Deep Set architecture for supervised WSI classification \citep{deepset_wsi}, we propose a novel combination of the Deep Set architecture and SimCLR. See Figure \ref{fig:ds_simclr_arch} for a depiction of this architecture.

Similarly to PS-SimCLR, we use $\omega$ to construct the window of a scan $S_i$ with start and end locations $j$ and $j + \omega L_i$. We sample $K$ equidistant images from this window: $\{x^{(i)}_l\}_{l=1}^K$. The augmentation functions within $\mathcal{A}$ are constructed using the same base transformations as in PS-SimCLR and the baseline, however, \emph{the random crop occurs at the same location for all $K$ sampled slices in the set}. This is done to ensure semantic consistency of the slices in the set. We then compute a random view $X_1$ of the scan: $X_1 \coloneqq \{\alpha_1(x^{(i)}_l)\}_{l=1}^K$ for $\alpha_1 \in \mathcal{A}$. We construct the other random view $X_2$ using a different random window of the scan. Note that for DS-SimCLR, random views are defined as a \textbf{set} of transformed slices, unlike PS-SimCLR or the baseline where views are individual transformed slices.

To compute latent representations for these set-based random views, we employ the Deep Set architecture. We project each transformed slice of $X_1$ and $X_2$ using $f$: $V_1 = \{f(x) | x \in X_1\}_{l=1}^K$ (similarly for $X_2$). Next, we sum the representations of the set's slices and send them through an MLP, formally (as per \cite{deepset}):
\begin{equation}
	\label{eqn:deepset}
	\tilde{h}_1 = g_{\texttt{DS}} \left( \sum_{j=1}^K V_{1, j} \right)
\end{equation}
where $g_{\texttt{DS}}$ is a function from the Deep Set architecture used to compute the representation of a set $\tilde{h}_1$. We do the same for $V_2$ to compute $\tilde{h}_2$. Finally, we compute the final latent representations from which we compute the NT-Xent loss: $z_1 = g(\tilde{h}_1)$ and $z_2 = g(\tilde{h}_2)$. Note that we experiment with different parameterisations of $g_{\texttt{DS}}$, including the identity function and an MLP.

The use of the DeepSet architecture means that slices of a scan are modelled in a permutation invariant way since we are not modelling the scan's 3D volume as a sequence, but instead as a set. This means we trade off the ability to model a scan as a sequence in favour of significantly lower computational demand. Further, Deep Sets have previously been shown as a promising way of modelling medical data \citep{deepset_wsi}. Note that we still do indeed incorporate the inherent 3D nature of the volume into our modelling process by treating it as a set. We ensure that subvolumes (along the axial plane) of the same scan are pulled together, and subvolumes from different scans are pushed apart. Note that if $K = 1$ and $g_{\texttt{DS}} = \texttt{id}$, our DS-SimCLR model is equivalent to the baseline model.

\subsection{Implementation Details}
\subsubsection{Datasets}
We pretrain on the NLST dataset, which consists of 191133 CT scans of human lungs. We use a random subset of 2000 scans to align with the size of pretraining datasets used in previous works \citep{2d3djointmedicalssl} and for computational efficiency reasons. For downstream evaluation, we use the PCAM, CRC, CovidX-CT, and CheXpert (henceforth denoted as CX) datasets for the linear evaluation task. We use the Kvasir Instruments and TBX11 datasets for downstream fine-tuning for the semantic segmentation (SemSeg) and object detection (OD) tasks, respectively. These datasets provide a variety of pathology-related tasks to ensure a comprehensive evaluation of our pretrained models.
\subsubsection{Architecture}
We parameterise $f$ as a ResNet50 \citep{resnet} and $g$ as the standard two-layer non-linear MLP in all cases. $g_{\texttt{DS}}$ is parameterised as a one-layer non-linear MLP. Note that we experiment with an alternate parameterisation of $g_{\texttt{DS}}$ as the identity function, denoted $\texttt{id}$.
\subsubsection{Optimisation}
In all cases, we train with an initial learning rate of $0.07$ (following the learning rate formula from \citep{simclr}) and implement cosine learning rate decay throughout training. We set the weight decay to $1e-10$, batch size to $32$, and train for $100$ epochs using the Adam \citep{adam} optimiser.
\subsubsection{Key Hyperparameters}
We train with an image size of $224 \times 224$ for the baseline and PS-SimCLR, and $128 \times 128$ for DS-SimCLR by default (to reduce computational complexity). For PS-SimCLR and DS-SimCLR, we set $\omega = 0.1$ and $\omega = 0.5$, respectively (unless otherwise specified). We set $T = 5$ in PS-SimCLR, and $K = 3$ by default in DS-SimCLR. Note that we experiment with other values for these hyperparameters in our analyses.

\section{Experiments}
\label{sec:experiments}

\subsection{Linear Evaluation on Downstream Datasets}
\begin{table*}[t]
\caption{Linear evaluation results of the baseline and two proposed variants, PS-SimCLR and DS-SimCLR, for various classification datasets.}
\label{tbl:linear_eval_results}
\centering
\begin{tabular}{l|c|c|c|c|c|c|}
\cline{2-7}
 & CRC & CovidX-CT & PCAM & CX (Atelectasis) & CX (Cardiomegaly) & CX (Edema) \\ \hline
\multicolumn{1}{|l|}{Standard} & 82.27 & 79.82 & 76.55 & 35.38 & 59.74 & 57.95 \\ \hline
\multicolumn{1}{|l|}{PS-SimCLR} & 83.69 & 81.96 & 79.22 & 34.79 & \pmb{62.91} & \pmb{61.71} \\ \hline
\multicolumn{1}{|l|}{DS-SimCLR} & \pmb{89.13} & \pmb{84.72} & \pmb{81.14} & \pmb{35.98} & 56.75 & 59.06 \\ \hline
\end{tabular}
\end{table*}

From Table \ref{tbl:linear_eval_results} we can see that our two proposed variants - PS-SimCLR and DS-SimCLR - consistently outperform the baseline for all linear evaluation datasets, with DS-SimCLR outperforming both by a large margin on the CRC, CovidX-CT, and PCAM datasets. For example, DS-SimCLR shows an 8.3\% and 6.1\% increase in performance for the CRC and CovidX-CT datasets compared to the baseline. Overall, DS-SimCLR shows a notable average relative increase in performance of 3.1\% across these datasets, and PS-SimCLR shows 3.0\% for the same metric. This suggests that implicitly encoding the 3D nature of medical datasets during pretraining assists with downstream linear evaluation performance, even for such tasks, which are not inherently 3D. This is also particularly notable because DS-SimCLR is trained on lower resolution random views than the baseline and PS-SimCLR in these experiments.

Our proposed methods are also fairly robust to the modality of the pathology in the downstream task. Pretraining was performed using the NLST datasets, which consists of 3D CT scans, and both variants perform well on downstream linear evaluation tasks consisting of X-rays (CheXpert), RGB histopathologic scans (CRC, PCAM), and CT scans (CovidX-CT). Interestingly, the results suggest that PS-SimCLR performs better than DS-SimCLR on 2 of the 3 studied CheXpert X-ray tasks. We posit that the stronger bias imposed upon DS-SimCLR regarding modelling the 3D nature of the pretraining data compared to PS-SimCLR's means that performance is better on this particular inherently 2D modality. It should be noted, however, that this does not reflect similarly in the inherently 2D tasks of CRC and PCAM. Finally, we note our improved performance within the linear evaluation paradigm suggests that the representations learned from our proposed methods better encode the domain-related information to make them more effective downstream as compared to the baseline. We posit this is mainly due to the implicit 3D modelling of PS-SimCLR's sampling strategy and DS-SimCLR's DeepSet component.

\subsection{Fine-Tuning on Downstream Datasets}
\begin{table}[!h]
\caption{Fine-tuning results of the baseline and two proposed variants, PS-SimCLR and DS-SimCLR, for the TBX11 OD task and Kvasir Instruments SemSeg task.}
\label{tbl:fine_tuning_results}
\centering
\begin{tabular}{l|c|c|}
\cline{2-3}
 & TBX11 & Kvasir Instruments \\ \hline
\multicolumn{1}{|l|}{Standard} & 23.04 & 78.99 \\ \hline
\multicolumn{1}{|l|}{PS-SimCLR} & 23.1 & 80.12 \\ \hline
\multicolumn{1}{|l|}{DS-SimCLR} & \pmb{23.37} & \pmb{81.03} \\ \hline
\end{tabular}
\end{table}

Table \ref{tbl:fine_tuning_results} shows the results of fine-tuning for two important medical tasks: OD and SemSeg. Note that in this setting, the encoder network $f$ changes its weights during training. Similar to the linear evaluation results shown in Table \ref{tbl:linear_eval_results}, both of our PS-SimCLR and DS-SimCLR proposed variants outperform the standard SSL baseline. The performance difference for these tasks is less pronounced as compared to linear evaluation. Apart from simply attributing this to a smaller performance difference, another possible reason is the sensitivity of the metric. The mAP and mIOU metrics are more sensitive to slight perturbations in predictions than accuracy. To investigate this further, we randomly sample 3 testing images and visualise their predicted segmentation masks.

\begin{figure*}
     \centering
     \begin{subfigure}[b]{0.3\textwidth}
         \centering
         \includegraphics[width=\textwidth]{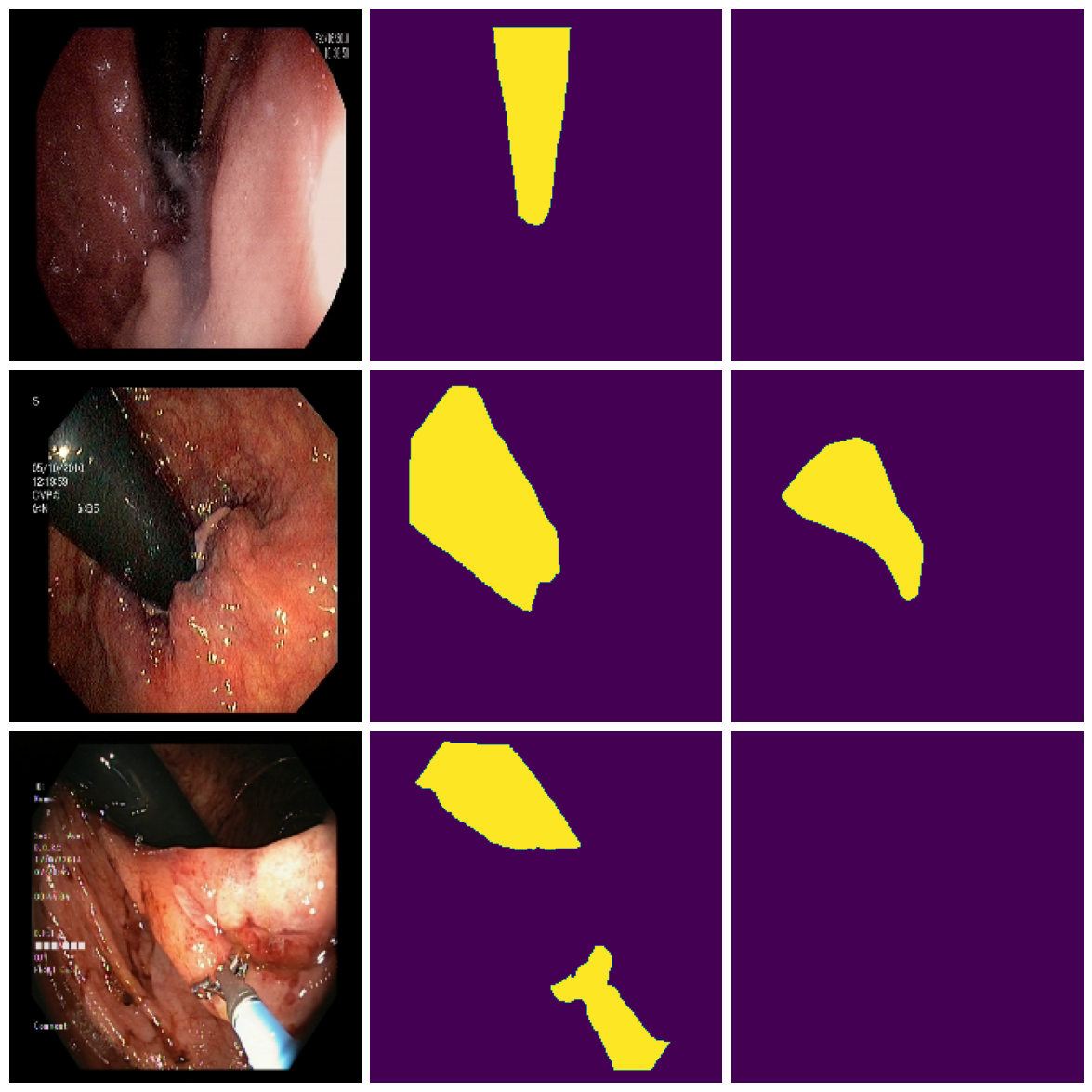}
         \caption{Standard.}
         \label{fig:standard_semseg_viz}
     \end{subfigure}
     \hfill
     \begin{subfigure}[b]{0.3\textwidth}
         \centering
         \includegraphics[width=\textwidth]{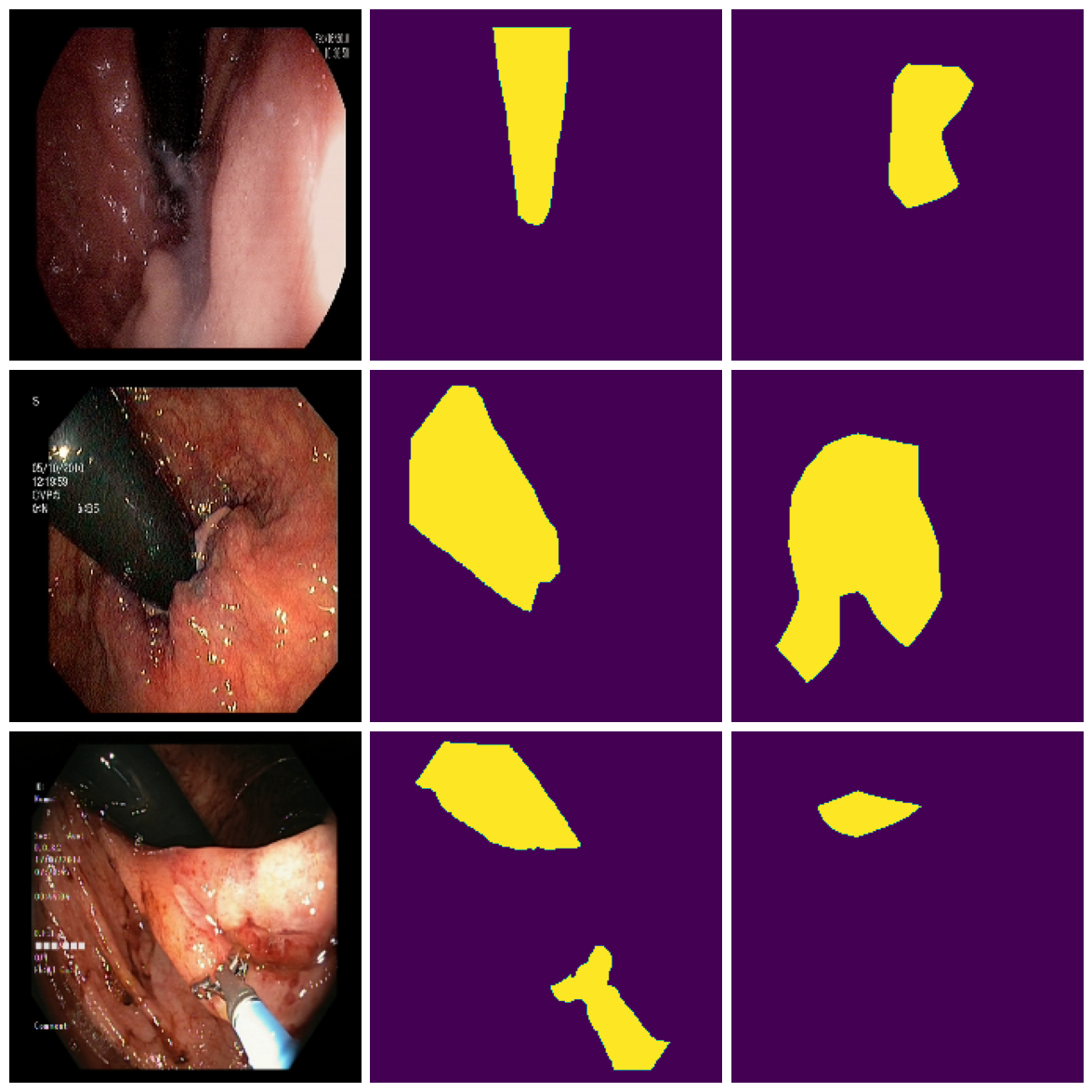}
         \caption{PS-SimCLR.}
         \label{fig:per_scan_semseg_viz}
     \end{subfigure}
     \hfill
     \begin{subfigure}[b]{0.3\textwidth}
         \centering
         \includegraphics[width=\textwidth]{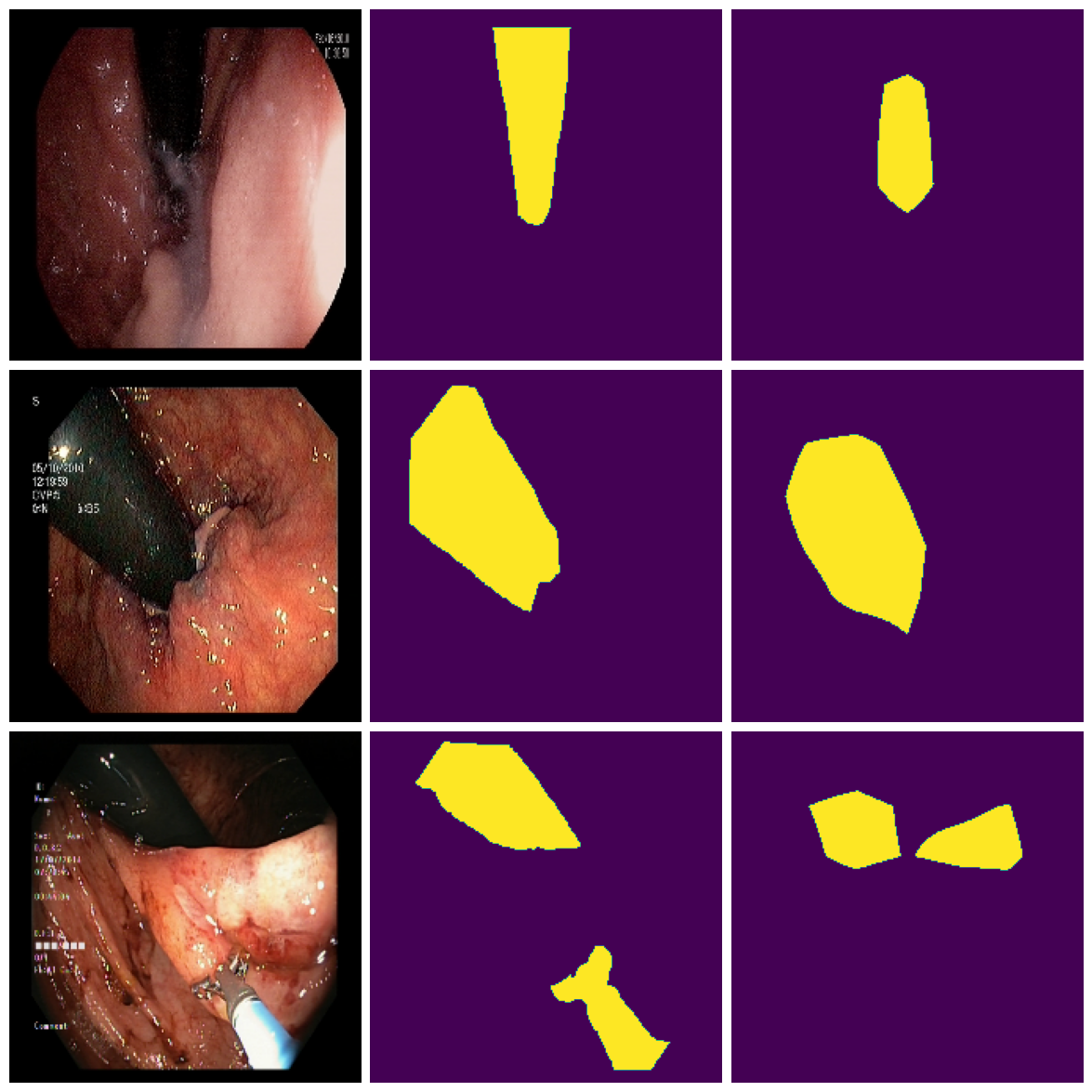}
         \caption{DS-SimCLR.}
         \label{fig:deepset_semseg_viz}
     \end{subfigure}
        \caption{Visualisation of SemSeg fine-tuning results on the Kvasir Instruments dataset for 3 randomly chosen test images. The three columns are L-R: input image, ground-truth mask, and predicted mask.}
        \label{fig:semseg_viz}
\end{figure*}

Figure \ref{fig:semseg_viz} visualises the SemSeg results for the 3 models. The predicted masks for DS-SimCLR are notably more coherent and accurate than both the baseline and PS-SimCLR. The standard model completely fails to detect the instrument in two of these random images. Further, PS-SimCLR is consistently able to detect that an instrument is present in the image, however, the localisations are not as accurate as DS-SimCLR. These results suggest that both proposed variants, and DS-SimCLR in particular, outperform the baseline in both sparse prediction tasks (linear evaluation) and these dense prediction tasks.

\subsection{Impact of Set Size and Sampling Width}
Next, we perform experiments to analyse how the most important hyperparameters introduced by two proposed methods affect performance. For PS-SimCLR, we focus on the sampling width $\omega$. We note that for PS-SimCLR, $\omega$ has a bigger impact on performance than $T$, and thus focus on it here. See Appendix \ref{tbl:ps_simclr_additional} for more experiments around hyperparameter $T$. For DS-SimCLR, we also analyse on sampling width $\omega$. Further, we include analyses of the set size $K$, as this is a core hyperparameter for the DeepSet architecture within our proposed framework.

\subsubsection{Set Size}
First, we analyse the impact of the set size $K$ on the performance of our DS-SimCLR model. We analyse three equally-spaced values for $K$ while keeping other hyperparameters constant. We use the values $K \in \{1, 3, 5\}$ to keep the computational complexity comparable with the baseline. Results can be seen in Table \ref{tbl:set_size_results}. When $K = 5$, linear evaluation (i.e. the classification datasets) performance is highest overall. However, the fine-tuning, dense prediction tasks of TBX11 and Kvasir Instruments result in the best performance for $K = 3$. Interestingly, $K = 1$ is consistently the worst-performing setting.

We hypothesise that the improved performance of the $K = 3$ setting versus the $K = 5$ setting for the fine-tuning, dense prediction tasks is because dense features may be harder to encode during learning when $K$ is larger since the variance in the set grows accordingly. When the variance is larger, two very distinct sets of images are encouraged to be similar through the use of the contrastive loss. Thus, detailed semantic information is lost through the aggregation of the distinct set of images. Similarly, when $K$ is too small (i.e. $K = 1$), we completely lose the benefit of the DeepSet and associated image set comparison, and therefore performance degrades.

\begin{table*}[!h]
\caption{Results of DS-SimCLR for various set sizes $K$. We set $\omega = 0.5$, $D = 64$, and $g_{\texttt{DS}} = \texttt{id}$ for these experiments. We sample three equidistant values for $K$.}
\label{tbl:set_size_results}
\centering
\begin{tabular}{l|c|c|c|c|c|c|c|c|}
\cline{2-9}
 & \multicolumn{1}{l|}{CRC} & \multicolumn{1}{l|}{CovidX-CT} & \multicolumn{1}{l|}{PCAM} & \multicolumn{1}{l|}{CX (Atelectasis)} & \multicolumn{1}{l|}{CX (Cardiomegaly)} & \multicolumn{1}{l|}{CX (Edema)} & \multicolumn{1}{l|}{TBX11} & \multicolumn{1}{l|}{Kvasir} \\ \hline
\multicolumn{1}{|l|}{$K = 1$} & 77.53 & 77.34 & 79.06 & 34.27 & 54.87 & 53.93 & 20.53 & 75.28 \\ \hline
\multicolumn{1}{|l|}{$K = 3$} & 88.15 & \pmb{84.6} & \pmb{80.83} & 34.7 & \pmb{60.09} & 57.61 & \pmb{22.19} & \pmb{78.56} \\ \hline
\multicolumn{1}{|l|}{$K = 5$} & \pmb{89.26} & 83.78 & 78.83 & \pmb{36.32} & 58.72 & \pmb{59.32} & 20.39 & 75.52 \\ \hline
\end{tabular}
\end{table*}

\subsubsection{Sampling Width}
The sampling width $\omega$ essentially controls the width of the contiguous window of the scan from which we can sample slices for PS-SimCLR and DS-SimCLR. Similar to the set size analysis, we vary $\omega$ while keeping all other hyperparameters constant. We test a small, medium, and large value for $\omega$ in both cases. Results for both variants can be seen in Table \ref{tbl:sampling_width_results}. We see that for PS-SimCLR, $\omega = 0.1$ results in the best performance overall, with that setting having the best performance in $6 / 8$ of the downstream tasks. In these two other tasks, the performance difference is small. We posit that this is because as $\omega \rightarrow 1$, the mutual information between the two sampled slices decreases. As the mutual information decreases, the semantic consistency between the two images decreases, which makes the contrastive task harder to solve.

For DS-SimCLR, we note that overall, $\omega = 0.5$ performs best (i.e. for $4 / 8$ tasks), however, there are 2 tasks each where $\omega = 0.8$ and $\omega = 0.2$ perform best. The reason there is no clear winner is because $\omega$ has less of an effect on the representations learned by the DeepSet architecture compared with the set size $K$ due to the equidistant sampling utilised to create the set. One can still achieve a balance of image variety and mutual information by sampling more images within the same sampling width region. It is clear from these results that $\omega = 0.5$ provides a good balance of this mutual information within the set such that useful representations can be learned during pretraining for a variety of downstream tasks.

\begin{table*}[!h]
\caption{Results of PS-SimCLR and DS-SimCLR for various sampling widths. For PS-SimCLR, we set $T = 5$. For DS-SimCLR, we set $K = 3$, $D = 64$, and $g_{\texttt{DS}} = \texttt{id}$. We sample three equidistant values for $\omega$ for each model.}
\label{tbl:sampling_width_results}
\centering
\resizebox{\textwidth}{!}{
\begin{tabular}{cc|c|c|c|c|c|c|c|c|}
\cline{3-10}
 &  & \multicolumn{1}{l|}{CRC} & \multicolumn{1}{l|}{CovidX-CT} & \multicolumn{1}{l|}{PCAM} & \multicolumn{1}{l|}{CX (Atelectasis)} & \multicolumn{1}{l|}{CX (Cardiomegaly)} & \multicolumn{1}{l|}{CX (Edema)} & \multicolumn{1}{l|}{TBX11} & \multicolumn{1}{l|}{Kvasir} \\ \hline
\multicolumn{1}{|l|}{\multirow{3}{*}{PS-SimCLR}} & $\omega = 0.1$ & \pmb{83.69} & \pmb{81.96} & \pmb{79.22} & 34.79 & 62.91 & \pmb{61.71} & \pmb{23.1} & \pmb{80.12} \\ \cline{2-10} 
\multicolumn{1}{|l|}{} & $\omega = 0.4$ & 79.72 & 78.26 & 76.79 & 34.96 & 61.03 & 54.7 & 22.32 & 79.64 \\ \cline{2-10} 
\multicolumn{1}{|l|}{} & $\omega = 0.7$ & 79.55 & 78.43 & 77.63 & \pmb{35.3} & \pmb{63.59} & 58.72 & 22.82 & 78.24 \\ \hline \hline
\multicolumn{1}{|l|}{\multirow{3}{*}{DS-SimCLR}} & $\omega = 0.8$ & 88.79 & 84.59 & \pmb{79.93} & 35.38 & 56.15 & 57.35 & \pmb{21.9} & 76.23 \\ \cline{2-10} 
\multicolumn{1}{|l|}{} & $\omega = 0.5$ & \pmb{89.26} & 83.78 & 78.83 & \pmb{36.32} & \pmb{58.72} & \pmb{59.32} & 20.39 & 75.52 \\ \cline{2-10} 
\multicolumn{1}{|l|}{} & $\omega = 0.2$ & 88.75 & \pmb{85.13} & 79.61 & 34.62 & 56.15 & 58.03 & 21.12 & \pmb{78.32} \\ \hline
\end{tabular}
}
\end{table*}

\subsection{Analysis of $g_{\texttt{DS}}$}
Next, we analyse the effect of different parameterisations of $g_{\texttt{DS}}$ in Equation \ref{eqn:deepset}. If we were to make $g_{\texttt{DS}}$ the identity function (i.e. $g_{\texttt{DS}}(x) = x$ $\forall x$), then we would leave it to the projection head $g$ to serve as the DeepSet's proxy MLP. This slightly reduces computational overhead by negating the need for a forward pass through a small MLP. Results for both the identity and MLP parameterisations of $g_{\texttt{DS}}$ can be seen in Table \ref{tab:deepset_mlp_results}.

The results suggest that for certain parameterisations of DS-SimCLR, there is a clear performance increase, on average, when an MLP is used (such as in $\texttt{DS-SimCLR}(3, 128, 0.5)$ and $\texttt{DS-SimCLR}(2, 224, 0.7)$). However, for different parameterisations, the identity function results in superior performance, such as in $\texttt{DS-SimCLR}(5, 64, 0.5)$ and $\texttt{DS-SimCLR}(5, 64, 0.8)$. These results suggest that the choice of $g_{\texttt{DS}}$ is heavily influenced by the choice of the remaining parameters $K$ $D$, and $\omega$. Interestingly, for the SemSeg downstream task (Kvasir), the best performance is always achieved when an MLP is used.

Moreover, if we aggregate the results by dataset based on $g_{\texttt{DS}}$'s parameterisation (see Table \ref{tbl:aggregated_g_ds_results} in the Appendix), we see that for the majority of downstream tasks, an MLP performs best on average. This suggests that decoupling the DeepSet's MLP from the projection head, and thus from the SimCLR architecture, results in better performance. However, we reiterate that this is largely dependent on the hyperparameter choices.

\begin{table*}[t]
\caption{Results for the inclusion or exclusion of an MLP in the DeepSet formulation (see Equation \ref{eqn:deepset}). Note that in $\texttt{DS-SimCLR}(K, D, \omega)$, $K$ is the set size, $D$ is the image size (height and width), and $\omega$ is the sampling width.}
\label{tab:deepset_mlp_results}
\centering
\resizebox{\textwidth}{!}{
\begin{tabular}{ll|cccccccc}
 & $g_{\texttt{DS}}$ & \multicolumn{1}{l}{CRC} & \multicolumn{1}{l}{CovidX-CT} & \multicolumn{1}{l}{PCAM} & \multicolumn{1}{l}{CX (Atelectasis)} & \multicolumn{1}{l}{CX (Cardiomegaly)} & \multicolumn{1}{l}{CX (Edema)} & \multicolumn{1}{l}{TBX11} & \multicolumn{1}{l}{Kvasir} \\ \hline
DS(5, 64, 0.5) & $\texttt{id}$ & \pmb{89.26} & 83.78 & \pmb{78.83} & \pmb{36.32} & \pmb{58.72} & \pmb{59.32} & \pmb{20.39} & 75.52 \\
DS(5, 64, 0.5) & MLP & 88.82 & \pmb{84.5} & 78.79 & 34.79 & 52.48 & 57.09 & 20.32 & \pmb{79.07} \\ \hline
DS(5, 64, 0.8) & $\texttt{id}$ & 88.79 & 84.59 & 79.93 & \pmb{35.38} & 56.15 & \pmb{57.35} & \pmb{21.9} & 76.23 \\
DS(5, 64, 0.8) & MLP & \pmb{88.92} & \pmb{85.56} & \pmb{80.55} & 34.7 & \pmb{60.68} & 55.81 & 21.09 & \pmb{77.28} \\ \hline
DS(3, 128, 0.5) & $\texttt{id}$ & 88.53 & 84.07 & \pmb{81.23} & 35.38 & \pmb{58.55} & 57.09 & 22.47 & 79.27 \\
DS(3, 128, 0.5) & MLP & \pmb{89.13} & \pmb{84.72} & 81.14 & \pmb{35.98} & 56.75 & \pmb{59.06} & \pmb{23.37} & \pmb{81.03} \\ \hline
DS(2, 224, 0.7) & $\texttt{id}$ & \pmb{83.08} & \pmb{82.91} & 78.58 & \pmb{35.13} & 62.05 & 56.5 & 20.4 & 75.73 \\
DS(2, 224, 0.7) & MLP & 82.82 & 81.61 & \pmb{79.35} & 35.04 & \pmb{64.62} & \pmb{60.0} & \pmb{22.65} & \pmb{78.17} \\ \hline
\end{tabular}
}
\end{table*}

\section{Conclusion}
\label{sec:conclusion}
We choose to model the 3D volumes as a set using the DeepSet since this architecture has proven to work well in supervised medical settings. Our integration of the DeepSet into an SS architecture (DS-SimCLR) has multiple benefits, including implicitly modelling the 3D nature of the data while notably improving performance for little additional overhead. In this way, we contrast against existing approaches which are either inherently 3D architectures or use complicated transformer-based mechanisms, both of which are significantly more computationally expensive to train than our proposed methods. Further, our simple sampling strategy proposed for PS-SimCLR improves upon the baseline for no additional training overhead.

\bibliographystyle{unsrtnat}
\bibliography{references}

\newpage
\appendix
\section{Additional PS-SimCLR Results}
\begin{table}[!h]
\caption{Additional results for different settings for $T$ for PS-SimCLR. It is clear that in both the $\omega = 0.1$ and $\omega = 0.7$ cases, there is no clear trend as to which value of $T$ is best. This suggests that the method, PS-SimCLR, is fairly robust to the choice of this parameter. However, we recommend the value be kept fairly small relative to the average number of slices in a scan for a particular dataset. This ensures that we do not revert to the baseline sampling during training often.}
\label{tbl:ps_simclr_additional}
\centering
\resizebox{\textwidth}{!}{
\begin{tabular}{ll|c|c|c|c|c|c|c|c|}
\cline{3-10}
 &  & \multicolumn{1}{l|}{CRC} & \multicolumn{1}{l|}{CovidX-CT} & \multicolumn{1}{l|}{PCAM} & \multicolumn{1}{l|}{CX (Atelectasis)} & \multicolumn{1}{l|}{CX (Cardiomegaly)} & \multicolumn{1}{l|}{CX (Edema)} & \multicolumn{1}{l|}{TBX11} & \multicolumn{1}{l|}{Kvasir} \\ \hline
\multicolumn{1}{|l|}{\multirow{2}{*}{$\omega = 0.1$}} & $T = 5$ & \pmb{83.69} & \pmb{81.96} & \pmb{79.22} & 34.79 & \pmb{62.91} & \pmb{61.71} & \pmb{23.1} & \pmb{80.12} \\ \cline{2-10} 
\multicolumn{1}{|l|}{} & $T = 50$ & 82.88 & 80.45 & 78.69 & \pmb{36.58} & 58.46 & 61.54 & 22.79 & 79.07 \\ \hline
\multicolumn{1}{|l|}{\multirow{2}{*}{$\omega = 0.7$}} & $T = 5$ & 79.55 & \pmb{78.43} & 77.63 & \pmb{35.3} & 63.59 & \pmb{58.72} & \pmb{22.82} & 78.24 \\ \cline{2-10} 
\multicolumn{1}{|l|}{} & $T = 50$ & \pmb{79.96} & 76.58 & \pmb{78.04} & 35.13 & \pmb{66.07} & 57.95 & 21.88 & \pmb{78.38} \\ \hline
\end{tabular}
}
\end{table}

\section{Aggregated $g_\texttt{DS}$ Results}
\begin{table}[!h]
\caption{Aggregated results of different parameterisations of $g_{\texttt{DS}}$.}
\label{tbl:aggregated_g_ds_results}
\centering
\resizebox{\textwidth}{!}{
\begin{tabular}{c|c|c|c|c|c|c|c|c|}
\cline{2-9}
 & CRC & CovidX-CT & PCAM & CX (Atelectasis) & CX (Cardiomegaly) & CX (Edema) & TBX11 & Kvasir \\ \hline
\multicolumn{1}{|l|}{\texttt{id}} & \pmb{87.42} & 83.84 & 79.64 & \pmb{35.55} & \pmb{58.87} & 57.56 & 21.29 & 76.69 \\ \hline
\multicolumn{1}{|l|}{MLP} & \pmb{87.42} & \pmb{84.1} & \pmb{79.96} & 35.13 & 58.63 & \pmb{57.99} & \pmb{21.86} & \pmb{78.89} \\ \hline
\end{tabular}
}
\end{table}

\end{document}